\documentclass[10pt,twocolumn,letterpaper]{article}

\usepackage{iccv}
\usepackage{amsmath,graphicx}
\usepackage{booktabs}
\usepackage{float}
\usepackage{xcolor}
\usepackage{subcaption} 

\definecolor{red}{RGB}{255, 0, 0}

\usepackage[breaklinks=true,bookmarks=false]{hyperref}

\iccvfinalcopy 

\def\iccvPaperID{****} 

\begin{document}

\title{SiamReID: Confuser Aware Siamese Tracker with Re-identification Feature}
\author{Abu Md Niamul Taufique$^{\star\dagger}$ \quad Andreas Savakis$^{\dagger}$ \quad Michael Braun$^{\ddagger}$ \quad Daniel Kubacki$^{\ddagger}$\\ \quad Ethan Dell$^{\ddagger}$ \quad Lei Qian$^{\mathsection}$ \quad Sean M. O'Rourke$^{\mathsection}$ \\$\star$ at7133@rit.edu\\ $\dagger$ Rochester Institute of Technology \\
$\ddagger$ Systems \& Technology Research \\
$\mathsection$ Air Force Research Labs \\
}




\maketitle

\begin{abstract}
Siamese deep-network trackers have received significant attention in recent years due to their real-time speed and state-of-the-art performance. However, Siamese trackers suffer from similar looking confusers, that are prevalent in aerial imagery and create challenging conditions due to prolonged occlusions where the tracker object re-appears under different pose and illumination. Our work proposes SiamReID, a novel re-identification framework for Siamese trackers, that incorporates confuser rejection during prolonged occlusions and is well-suited for aerial tracking. The re-identification feature is trained using both triplet loss and a class balanced loss. Our approach achieves state-of-the-art performance in the UAVDT single object tracking benchmark.
\end{abstract}

\section{Introduction}
\label{sec:intro}




Deep learning-based visual object tracking methods, such as tracking by detection \cite{nam2016learning}, correlation filters \cite{danelljan2017eco} and Siamese trackers \cite{bertinetto2016fully, li2018high}, have been widely used due to their superior performance.
Aerial videos bring a unique set of challenges to tracking, such as camera motion and rotation, small object size, long-term occlusion, out-of-view movement, etc. \cite{taufique2020benchmarking}.
Results by Taufique et al. \cite{taufique2020benchmarking} showed that even though state-of-the-art trackers perform well on ground-level datasets, their performance significantly degrades in aerial datasets.
In this paper, we incorporate a re-identification feature to Siamese tracking for improved performance in aerial videos.

In recent years, Siamese-based and correlation filter-based trackers are the most popular types of trackers
because of their performance and potential for real-time operation. Bertinetto et al. \cite{bertinetto2016fully} introduced a Siamese tracking framework where a response map is computed based on the cross-correlation of the extracted features from a template frame with a search frame. Li et al. \cite{li2018high} extended the idea of Siamese tracking with a Region Proposal Network (RPN) in SiamRPN, which significantly improves the bounding box estimation accuracy. Recently, Wang et al. \cite{wang2019fast} proposed the SiamMask tracker to take advantage of a deeper backbone with depth-wise cross-correlation that also significantly improved tracking performance.

\begin{figure}[t]

\begin{minipage}[b]{1.0\linewidth}
  \centering
  \centerline{\includegraphics[width=5cm,height=5cm]{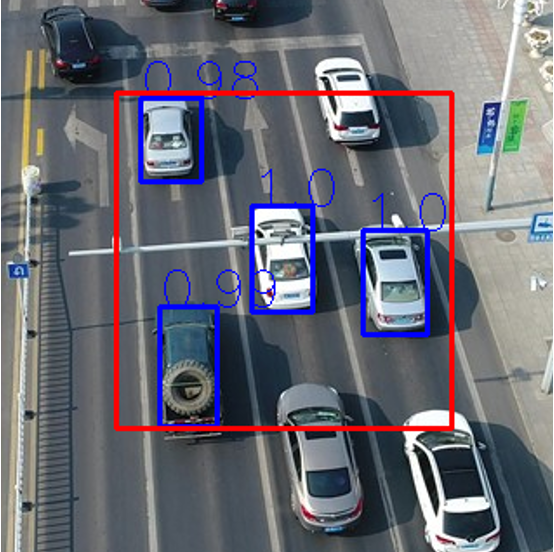}}
\end{minipage}
\caption{Siamese network classification scores for vehicles (blue boxes) inside the search region (red box) are similar despite their appearance differences.}
\label{fig:siamese_score}
\end{figure}

Despite their successes, Siamese trackers have limitations  under occlusion or out-of-frame conditions
due to similar looking confusers, also known as distractors.
\cite{taufique2020benchmarking, bhat2019learning}.
As illustrated in Fig. \ref{fig:siamese_score},
the Siamese network is
not effective in separating individual instances of the same object class.
The cosine windowing technique with location information is generally used to avoid jumping to confusers. However, during long-term occlusions or out-of-view situations, the cosine window is no longer applicable, and the tracked object re-acquisition needs to be purely appearance based.

We propose SiamReID, a two-stage network that improves Siamese
object tracking,
where a detection algorithm locates objects in a sequence of frames and a re-identification (Re-ID) algorithm associates these detections for re-acquisition \cite{yu2016poi}. The Siam-ReID framework presented here is tailored to vehicle tracking in aerial imagery and
extends the capabilities of the SiamRPN tracker family \cite{li2018high}.
In a given frame, we select the bounding boxes with high score predictions by the Siamese network. Then, we apply the Re-ID algorithm to separate the confusers from the target. When the target is lost, e.g. due to occlusion, we dynamically increase the search area until the Re-ID algorithm re-acquires the target. In short, the Re-ID algorithm acts as an oracle for confuser rejection and re-identification.
Our main contributions are as follows.
\begin{enumerate}
\vspace{-0.05in}
    \item We present a two-stage tracking framework with re-identification that improves the robustness of the Siamese visual object tracking framework under occlusion or out-of-view conditions.
\vspace{-0.1in}
    \item We propose a vehicle Re-ID feature that incorporates confuser rejection and target re-identification capabilities in Siamese trackers.
\vspace{-0.1in}
    \item Our proposed SiamReID tracker  outperforms the state-of-the-art trackers on UAVDT vehicle tracking benchmark.
\vspace{-0.1in}
\end{enumerate}

\section{Related Works}
\label{sec:relatedworks}

\subsection{Visual object tracking}

After the introduction of the Siamese fully-convolutional tracker by SiamFC \cite{bertinetto2016fully}, SiamRPN \cite{li2018high} and SiamMask \cite{wang2019fast} were notable improvements that incorporated  bounding box detection and semantic segmentation capabilities respectively.
Furthermore, SiamRPN++ \cite{li2019siamrpn++} utilized a deeper network compared to SiamRPN \cite{li2018high} with depth-wise cross-correlation, and achieved better performance.
SiamRPN++ utilized intermediate features from the backbone network for the final output, which showed improvement.

Recently, Yang et al. \cite{yang2020release} proposed an online training strategy to improve tracking by developing a novel loss function to increase the inter-class distance while reducing the intra-class variance between target and background classes.
Li et al. \cite{li2020autotrack} proposed Autotrack that automatically learns the spatio-temporal regularization parameters for correlation filter based tracking.
Fu et al. \cite{fu2020disruptor} proposed mitigating correlation filter response map inconsistency by incorporating historical interval information.
Lin et al. \cite{lin2020learning} proposed a correlation filter based tracking framework that learns bi-directional tracking to improve generalizability of the tracker for unexpected appearance change.

Videos taken from an aerial perspective introduce additional challenges such as camera movement and lower resolution \cite{taufique2020benchmarking}. Minnehan et al. \cite{minnehan2019fully} proposed an adaptive template to learn the exemplar for object tracking from an aerial perspective.
In our work, we take advantage of the RPN \cite{ren2015faster} capabilities in Siamese trackers, such as SiamRPN \cite{li2018high}, that generally provide accurate bounding box and utilize a Re-ID model to handle the confusers and target re-acquisition.

\subsection{Vehicle re-identification}
In vehicle re-identification, the objective is to recognize a specific vehicle under various poses, illuminations, partial occlusion, etc., among a set of vehicles. He et al. \cite{he2020multi} proposed a strong baseline for vehicle Re-ID. Taufique et al. \cite{taufique2020LABNet} proposed to use a class-balanced loss \cite{cui2019class} and graph neural network that showed improved vehicle re-identification performance over the baseline.  In the Siamese tracking framework, we propose to use a simpler version of the Re-ID feature without the graph network.

\section{Siamese Tracking with Re-ID Methodology}
\label{sec:method}

Our framework consists of a Siamese network that acts as a base tracker and a Re-ID algorithm which rejects confusers and performs re-identification. In this paper, the Re-ID feature is trained for tracking vehicles in aerial videos.

\begin{figure}[htb]

\begin{minipage}[b]{1.0\linewidth}
  \centering
  \centerline{\includegraphics[width=\textwidth]{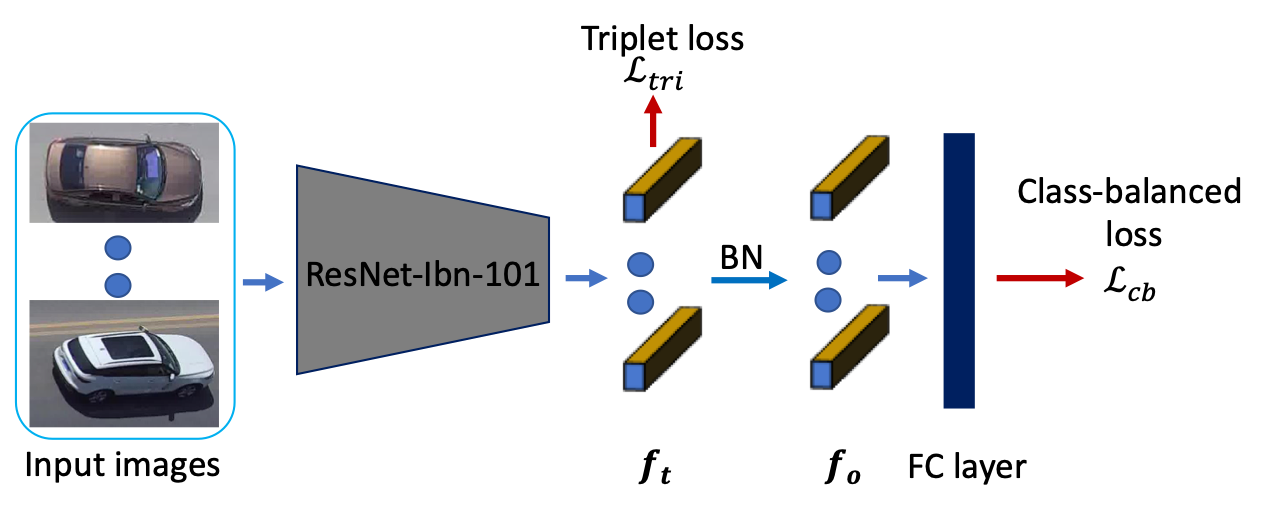}}
\end{minipage}
\caption{Vehicle re-identification network.}
\label{fig:reid_network}
\end{figure}

\begin{figure*}[htb]

\begin{minipage}[b]{1.0\linewidth}
  \centering
  \centerline{\includegraphics[width=\textwidth]{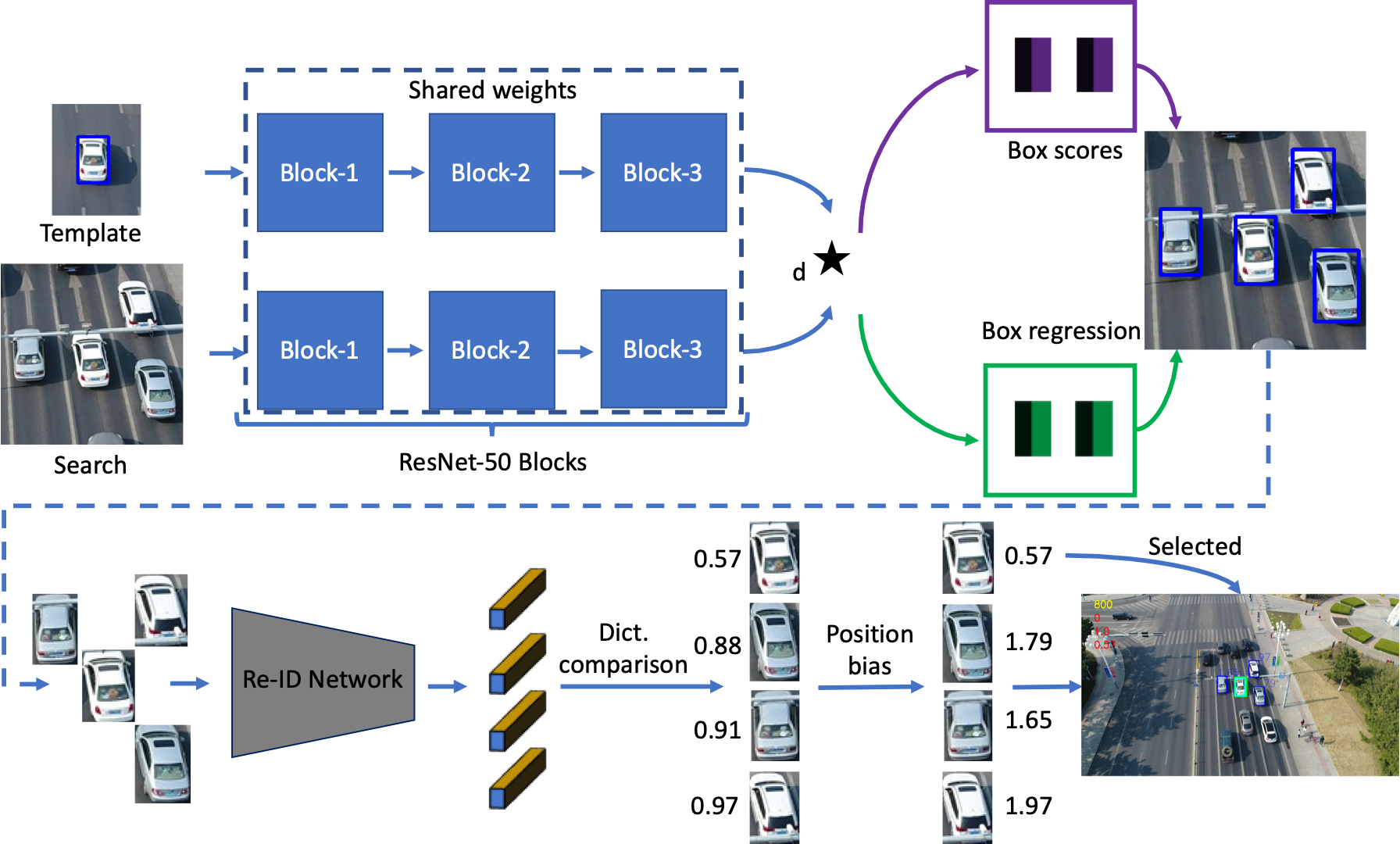}}
\end{minipage}
\caption{Overall tracking framework. Siamese tracker performs detection, and the re-identification network rejects the confusers. The position bias in the re-identification network is not applied if the target is lost.}
\label{fig:network}
\end{figure*}

\subsection{Re-identification Network}
Our Re-ID network, shown in Figure \ref{fig:reid_network}, is inspired from \cite{he2020multi, taufique2020LABNet}.
Given an input image $x \in \mathbf{R}^{W\times H \times 3}$, where, $W$ and $H$ are the width and the height of the image, respectively, the backbone $(BB)$ extracts the features followed by a Global Average Pooling (GAP) layer $G$. The generated feature, $f_t \in \mathbf{R}^{2048}$, then passes through a Batch Normalization (BN) neck $BN$ that is followed by the classifier. Finally, we use the output feature vector $f_o = BN(G(BB(x)))$, $f_o \in \mathbf{R}^{2048}$ to compute the cosine distance for re-identification.

\subsection{Tracking with re-identification}
The overall workflow of our method is depicted in Figure \ref{fig:network}.
We start with
a Siamese tracker, e.g. SiamRPN,
with the object template patch given in the first frame, and for subsequent frames we crop search patches centered on the previous location of the tracked object.
The extracted template and the search patch are depth-wise cross-correlated in the feature space.
The correlation map is  passed onto the ``Box scores" head and the ``Box regression" head, where the classification scores associated with each of the anchor boxes are predicted.
We perform a thresholding operation on the classification scores to
eliminate low confidence
bounding boxes from the total of $25\times25\times5$ predicted anchor boxes. We map the remaining anchor boxes to the image space to get the predictions.
Such predictions are shown in Figure \ref{fig:network}.

Let us assume that we get $N$ number of candidate bounding boxes. We perform online cropping of these candidate boxes and extract the Re-ID features $f_{o_i}$, where $i\in N$ for each of the objects. These features are compared with a dynamic dictionary representative, $f_{o_r}$, where the dynamic dictionary is constructed with the Re-ID features of all of the target crops that were tracked. We empirically found that the mean feature of the dictionary gives the best performance. We save the target re-id features at specific frame gaps to the dynamic dictionary. We compute the cosine distance for all of the candidate patches to the dictionary representative feature, which gives us the $appearance$ distance set $\mathcal{D}_a$
as follows.
\begin{equation}
    \mathcal{D}_a = \{Cos(f_{o_i}, f_{o_r})\}_{i=1}^{N}
\end{equation}

We additionally compute the Euclidean distance from the tracked center location of the bounding box at $(t-1)^{th}$ frame to all the candidate patches at $t^{th}$ which gives us the $position$ distance set $\mathcal{D}_e$, that can be written as follows.
\begin{equation}
    \mathcal{D}_e = \{Euclid(C^{(t-1)}, C^t_i)\}_{i=1}^{N},t\in T
\end{equation}
Here, $T$ is the total number of frames in a sequence, and $C$ is the center coordinate of a bounding box.
In the presence of confusers, we add a positional bias with the Re-ID distance that can be written as follows.
\begin{equation}
    d_i = {d_a}_i + \frac{d_{e_i} - min(\mathcal{D}_e)}{max(\mathcal{D}_e) - min(\mathcal{D}_e) + \epsilon}
\end{equation}
where $\epsilon$ is an arbitrary positive small number. The positional bias is added to minimize the impact of noisy estimation of the Re-ID model.
However, it is noteworthy that when the target is lost, we increase the search area and do not use any positional bias and only use appearance for re-acquisition.


\section{Dataset and Experiments}
\label{sec:dataset}
In our experiments, we use the UAVDT \cite{du2018unmanned} dataset for our tracker performance evaluation and VRAI \cite{wang2019vehicle} dataset for our Re-ID feature training.
The \textbf{UAVDT} \cite{du2018unmanned} dataset consists of 50 sequences where the camera is mounted with a drone and the targets are vehicles. There are several challenges, such as background clutter, occlusion, camera motion, significant scale change, etc.
The \textbf{VRAI} \cite{wang2019vehicle}
proposed a unique dataset for vehicle re-identification from an aerial perspective.


We performed data augmentation on VRAI with random  erasing, scaling and cropping
to train the Re-ID model. We also used $random identity$ sampler to sample 6 instances of 8 vehicles at each batch, e.g., batch size of 48. During training we linearly increased the learning rate from $10^{-3}$ to $10^{-2}$ in 10 epochs. The learning rate is reduced to $10^{-3}$ at epoch 40 and again reduced to $10^{-4}$ at epoch 70. The overall network is trained until epoch 120.

The Siamese tracker is initialized with the SiamMask \cite{wang2019fast} pretrained network weights, where we removed the mask branch to utilize the bounding box prediction only.
\begin{figure}[htb]
\centering
\subcaptionbox{Success plot.}{\includegraphics[width=0.45\textwidth]{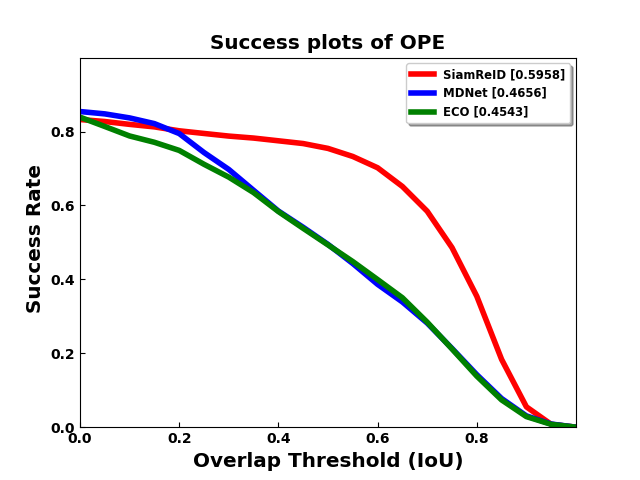}}%
\\ 
\subcaptionbox{Precision plot.}{\includegraphics[width=0.45\textwidth]{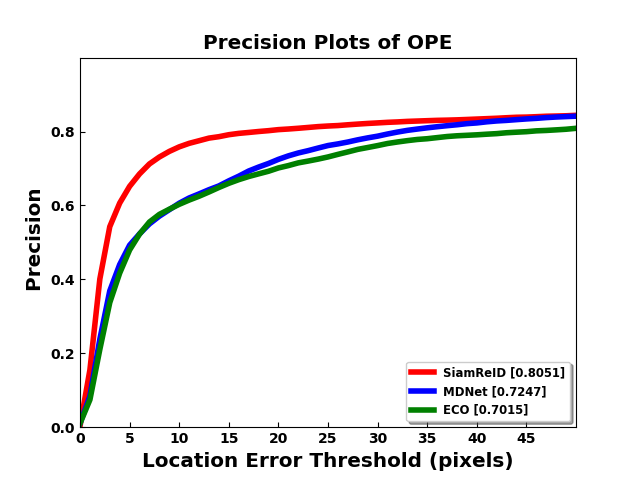}}%
\caption{One Pass Evaluation (OPE) results on UAVDT dataset.}
\label{fig:accuracy}
\end{figure}
\begin{figure}[htb]
\centering
\begin{minipage}[b]{0.85\linewidth}
  \centering
  \centerline{\includegraphics[width=\textwidth]{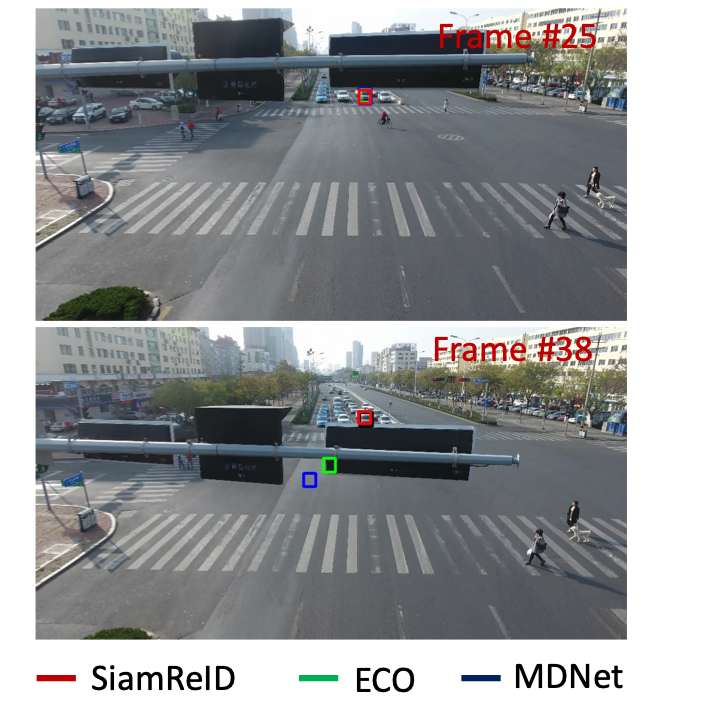}}
\end{minipage}
\caption{Visual results on UAVDT dataset showing two frames
from the sequence ``S0601". All three trackers successfully track the target before the full occlusion due to camera motion. Only SiamReID is able to re-acquire the target (red box).}
\label{fig:seqs}
\end{figure}

\section{Results and Discussion}
\label{sec:results}

\begin{table}[hb]
\caption{Results on VERI dataset}
\label{tab:performance}
\centering
\setlength\tabcolsep{1.0pt}
{\begin{tabular}{lccr}
\toprule
Method & Accuracy & Robustness & Venue \\
\midrule
SiamReID & 0.596 & 0.805 & Proposed \\
RPOT\cite{yang2020release} & 0.480 & 0.742 & AAAI 2020 \\
AutoTrack\cite{li2020autotrack} & 0.450 & 0.718 & CVPR 2020 \\
IBRI\cite{fu2020disruptor} & 0.470 & 0.738 & TGARS 2020 \\
TB-BiCF\cite{lin2020learning} & 0.468 & 0.727 & TCSV 2020 \\
\midrule
SiamRPN\cite{li2018high} & 0.565 & 0.74 & CVPR 2018 \\
\midrule
ECO\cite{danelljan2017eco} & 0.454 & 0.702 & CVPR 2017 \\
\midrule
MDNet\cite{nam2016learning} & 0.466 & 0.725 & CVPR 2016 \\

\bottomrule
\end{tabular}}
\end{table}

The comparison of SiamReID with state-of-the-art (SOTA) trackers is shown in Table \ref{tab:performance}.
The results show that
SiamReID outperforms several SOTA trackers.
It outperforms RPOT \cite{yang2020release}
by a margin of more than 10\% in accuracy 6\% in robustness and AutoTrack \cite{li2020autotrack} by more than 13\% and 8\% in accuracy and robustness, respectively.
We also compare our method with MDNet \cite{nam2016learning} and ECO \cite{danelljan2017eco} trackers on draw the accuracy and precision plot of OPE, which is shown in Figure \ref{fig:accuracy}. Our method outperforms these trackers by a significant margin.
We also visually compare our results with other methods and show an example
in Figure \ref{fig:seqs}.
In the S0601 video, both ECO and MDNet lost the target after full occlusion but SiamReID was able to re-acquire the target.

\section{Conclusion}
\label{sec:conclusion}
We proposed a novel two-stage tracking framework based on a Siamese network for short-term tracking and a re-identification algorithm for confuser rejection and target re-acquisition after full occlusion. We evaluated our method on an aerial vehicle tracking benchmark and achieved state-of-the-art performance.

\section*{Acknowledgements}
This research was supported in part by the Air Force Research Laboratory, Sensors Directorate (AFRL/RYAP) under a contract to Systems and Technology Research and an AFOSR grant. The authors acknowledge the computational resources made available by Research Computing at Rochester Institute of Technology that helped produce part of the results.

{\small
\bibliographystyle{ieee}
\bibliography{egbib}

\begin{thebibliography}{10}\itemsep=-1pt

\bibitem{bertinetto2016fully}
L.~Bertinetto, J.~Valmadre, J.~F. Henriques, A.~Vedaldi, and P.~H. Torr.
\newblock Fully-convolutional siamese networks for object tracking.
\newblock In {\em European Conference on Computer vision (ECCV)}, pages
  850--865. Springer, 2016.

\bibitem{bhat2019learning}
G.~Bhat, M.~Danelljan, L.~V. Gool, and R.~Timofte.
\newblock Learning discriminative model prediction for tracking.
\newblock In {\em IEEE International Conference on Computer Vision (ICCV)},
  pages 6182--6191, 2019.

\bibitem{cui2019class}
Y.~Cui, M.~Jia, T.-Y. Lin, Y.~Song, and S.~Belongie.
\newblock Class-balanced loss based on effective number of samples.
\newblock In {\em IEEE Conference on Computer Vision and Pattern Recognition
  (CVPR)}, pages 9268--9277, 2019.

\bibitem{danelljan2017eco}
M.~Danelljan, G.~Bhat, F.~Shahbaz~Khan, and M.~Felsberg.
\newblock {ECO}: Efficient convolution operators for tracking.
\newblock In {\em IEEE Conference on Computer Vision and Pattern Recognition
  (CVPR)}, pages 6638--6646, 2017.

\bibitem{du2018unmanned}
D.~Du, Y.~Qi, H.~Yu, Y.~Yang, K.~Duan, G.~Li, W.~Zhang, Q.~Huang, and Q.~Tian.
\newblock The unmanned aerial vehicle benchmark: Object detection and tracking.
\newblock In {\em European Conference on Computer Vision (ECCV)}, pages
  370--386, 2018.

\bibitem{fu2020disruptor}
C.~Fu, J.~Ye, J.~Xu, Y.~He, and F.~Lin.
\newblock Disruptor-aware interval-based response inconsistency for correlation
  filters in real-time aerial tracking.
\newblock {\em IEEE Transactions on Geoscience and Remote Sensing}, 2020.

\bibitem{he2020multi}
S.~He, H.~Luo, W.~Chen, M.~Zhang, Y.~Zhang, F.~Wang, H.~Li, and W.~Jiang.
\newblock Multi-domain learning and identity mining for vehicle
  re-identification.
\newblock In {\em IEEE Conference on Computer Vision and Pattern Recognition
  Workshops}, pages 582--583, 2020.

\bibitem{li2019siamrpn++}
B.~Li, W.~Wu, Q.~Wang, F.~Zhang, J.~Xing, and J.~Yan.
\newblock Siamrpn++: Evolution of siamese visual tracking with very deep
  networks.
\newblock In {\em IEEE Conference on Computer Vision and Pattern Recognition
  (CVPR)}, pages 4282--4291, 2019.

\bibitem{li2018high}
B.~Li, J.~Yan, W.~Wu, Z.~Zhu, and X.~Hu.
\newblock High performance visual tracking with siamese region proposal
  network.
\newblock In {\em IEEE Conference on Computer Vision and Pattern Recognition
  (CVPR)}, pages 8971--8980, 2018.

\bibitem{li2020autotrack}
Y.~Li, C.~Fu, F.~Ding, Z.~Huang, and G.~Lu.
\newblock Autotrack: Towards high-performance visual tracking for uav with
  automatic spatio-temporal regularization.
\newblock In {\em IEEE Conference on Computer Vision and Pattern Recognition},
  pages 11923--11932, 2020.

\bibitem{lin2020learning}
F.~Lin, C.~Fu, Y.~He, F.~Guo, and Q.~Tang.
\newblock Learning temporary block-based bidirectional incongruity-aware
  correlation filters for efficient uav object tracking.
\newblock {\em IEEE Transactions on Circuits and Systems for Video Technology},
  2020.

\bibitem{minnehan2019fully}
B.~Minnehan, A.~M.~N. Taufique, and A.~Savakis.
\newblock Fully convolutional adaptive tracker with real time performance.
\newblock In {\em SPIE Geospatial Informatics IX}, volume 10992, page 1099204,
  2019.

\bibitem{nam2016learning}
H.~Nam and B.~Han.
\newblock Learning multi-domain convolutional neural networks for visual
  tracking.
\newblock In {\em IEEE Conference on Computer Vision and Pattern Recognition
  (CVPR)}, pages 4293--4302, 2016.

\bibitem{ren2015faster}
S.~Ren, K.~He, R.~Girshick, and J.~Sun.
\newblock Faster r-cnn: Towards real-time object detection with region proposal
  networks.
\newblock In {\em Advances in Neural Information Processing Systems}, pages
  91--99, 2015.

\bibitem{taufique2020benchmarking}
A.~M.~N. Taufique, B.~Minnehan, and A.~Savakis.
\newblock Benchmarking deep trackers on aerial videos.
\newblock {\em Sensors}, 20(2):547, 2020.

\bibitem{taufique2020LABNet}
A.~M.~N. Taufique and A.~Savakis.
\newblock {LABNet}: Local graph aggregation network with class balanced loss
  for vehicle re-identification.
\newblock {\em arXiv preprint arXiv:2011.14417}, 2020.

\bibitem{wang2019vehicle}
P.~Wang, B.~Jiao, L.~Yang, Y.~Yang, S.~Zhang, W.~Wei, and Y.~Zhang.
\newblock Vehicle re-identification in aerial imagery: Dataset and approach.
\newblock In {\em IEEE International Conference on Computer Vision (ICCV)},
  pages 460--469, 2019.

\bibitem{wang2019fast}
Q.~Wang, L.~Zhang, L.~Bertinetto, W.~Hu, and P.~H. Torr.
\newblock Fast online object tracking and segmentation: A unifying approach.
\newblock In {\em IEEE Conference on Computer Vision and Pattern Recognition
  (CVPR)}, pages 1328--1338, 2019.

\bibitem{yang2020release}
Y.~Yang, G.~Li, Y.~Qi, and Q.~Huang.
\newblock Release the power of online-training for robust visual tracking.
\newblock In {\em AAAI Conference on Artificial Intelligence}, volume~34, pages
  12645--12652, 2020.

\bibitem{yu2016poi}
F.~Yu, W.~Li, Q.~Li, Y.~Liu, X.~Shi, and J.~Yan.
\newblock Poi: Multiple object tracking with high performance detection and
  appearance feature.
\newblock In {\em European Conference on Computer Vision (ECCV)}, pages 36--42.
  Springer, 2016.

\end{thebibliography}
}

\end{document}